\documentclass[10pt,twocolumn,letterpaper]{article}

\usepackage[pagenumbers]{cvpr} % To force page numbers, e.g. for an arXiv version

\usepackage{graphicx}
\usepackage{amsmath}
\usepackage[accsupp]{axessibility}
\usepackage{amssymb}
\usepackage{booktabs}
\usepackage[noend]{algpseudocode}

\usepackage{algorithmicx,algorithm}

\usepackage{times}
\usepackage{epsfig}
\usepackage{pifont}       % \ding{xx}
\usepackage{bbding}       % \Checkmark,\XSolid,... (需要和pifont宏包共同使用)
\usepackage{setspace}
\usepackage{multirow}
\usepackage{cite}
\usepackage{comment}
\usepackage{booktabs}
\usepackage{arydshln}
\usepackage{color}
\usepackage{xcolor}
\usepackage{bm}
\usepackage[normalem]{ulem}

\usepackage[pagebackref,breaklinks,colorlinks]{hyperref}

\usepackage[capitalize]{cleveref}
\crefname{section}{Sec.}{Secs.}
\Crefname{section}{Section}{Sections}
\Crefname{table}{Table}{Tables}
\crefname{table}{Tab.}{Tabs.}

\begin{document}

%%%%%%%%% TITLE - PLEASE UPDATE
\title{I$^2$CANSAY: Inter-Class Analogical Augmentation and Intra-Class Significance Analysis for Non-Exemplar Online Task-Free Continual Learning}

\author{Songlin Dong\textsuperscript{\rm 1 2\#}, Yingjie Chen\textsuperscript{\rm 1\#}, Yuhang He\textsuperscript{\rm 1}{\thanks{\* Corresponding authors,  \# Songlin Dong and Yingjie Chen are co-first authors}}, Yuhan Jin\textsuperscript{\rm 1}, Alex C. Kot\textsuperscript{\rm 2}, Yihong Gong\textsuperscript{\rm 1}\\ 
\textsuperscript{\rm 1} Xi'an Jiaotong University\\
\textsuperscript{\rm 2} Nanyang Technological University\\
% \textsuperscript{\rm 1}Xi'an Jiaotong University \quad
% \textsuperscript{\rm 2}Huawei Technologies Co., Ltd\\
% No.28, Xianning West Road, Xi'an, Shaanxi, 710049, P.R. China \\
%Xi'an, Shaanxi, China\\
{\tt\small \{dsl972731417,cyj1450670826,3122358042\}@stu.xjtu.edu.cn} \\
{\tt\small heyuhang@xjtu.edu.cn}, {\tt\small eackot@ntu.edu.sg}, {\tt\small ygong@mail.xjtu.edu.cn}}
\maketitle
\begin{abstract}
  Online task-free continual learning (OTFCL) is a more challenging variant of continual learning which emphasizes the gradual shift of task boundaries and learns in an online mode. Existing methods rely on a memory buffer composed of old samples to prevent forgetting. However, the use of memory buffers not only raises privacy concerns but also hinders the efficient learning of new samples. To address this problem, we propose a novel framework called I$^2$CANSAY that gets rid of the dependence on memory buffers and efficiently learns the knowledge of new data from one-shot samples.
  Concretely, our framework comprises two main modules. Firstly, the \textbf{Inter-Class Analogical Augmentation}~(ICAN) module generates diverse pseudo-features for old classes based on the inter-class analogy of feature distributions for different new classes, serving as a substitute for the memory buffer. Secondly, the \textbf{Intra-Class Significance Analysis}~(ISAY) module analyzes the significance of attributes for each class via its distribution standard deviation, and generates the importance vector as a correction bias for the linear classifier, thereby enhancing the capability of learning from new samples. We run our experiments on four popular image classification datasets: CoRe50, CIFAR-10, CIFAR-100, and CUB-200, our approach outperforms the prior state-of-the-art by a large margin. 
\end{abstract}

\section{Introduction}
Continual Learning (CL)~\cite{mai2022online,dong2021few,l2p} aims to incrementally learn knowledge from newly given data without forgetting the knowledge of old ones, and has drawn remarkable research interest in recent years. To date, most existing CL methods manually split a training dataset into a series of subsets (namely sessions or tasks), and then sequentially and incrementally train a unified model across these subsets. However, in real-world applications, the inputs to a CL system are often data streams (such as abnormal detection in a factory production line), where each sample is one-shot provided and the input data stream is seamless. This problem is referred to as the \textit{Online Task-Free Continual Learning} (OTFCL)~\cite{taskfree}, which is more challenging but practical.

\begin{figure}[tb]
\setlength{\abovecaptionskip}{-0.1cm} 
\setlength{\belowcaptionskip}{-0.4cm} 
\begin{center}
    \includegraphics[width=0.5\textwidth]{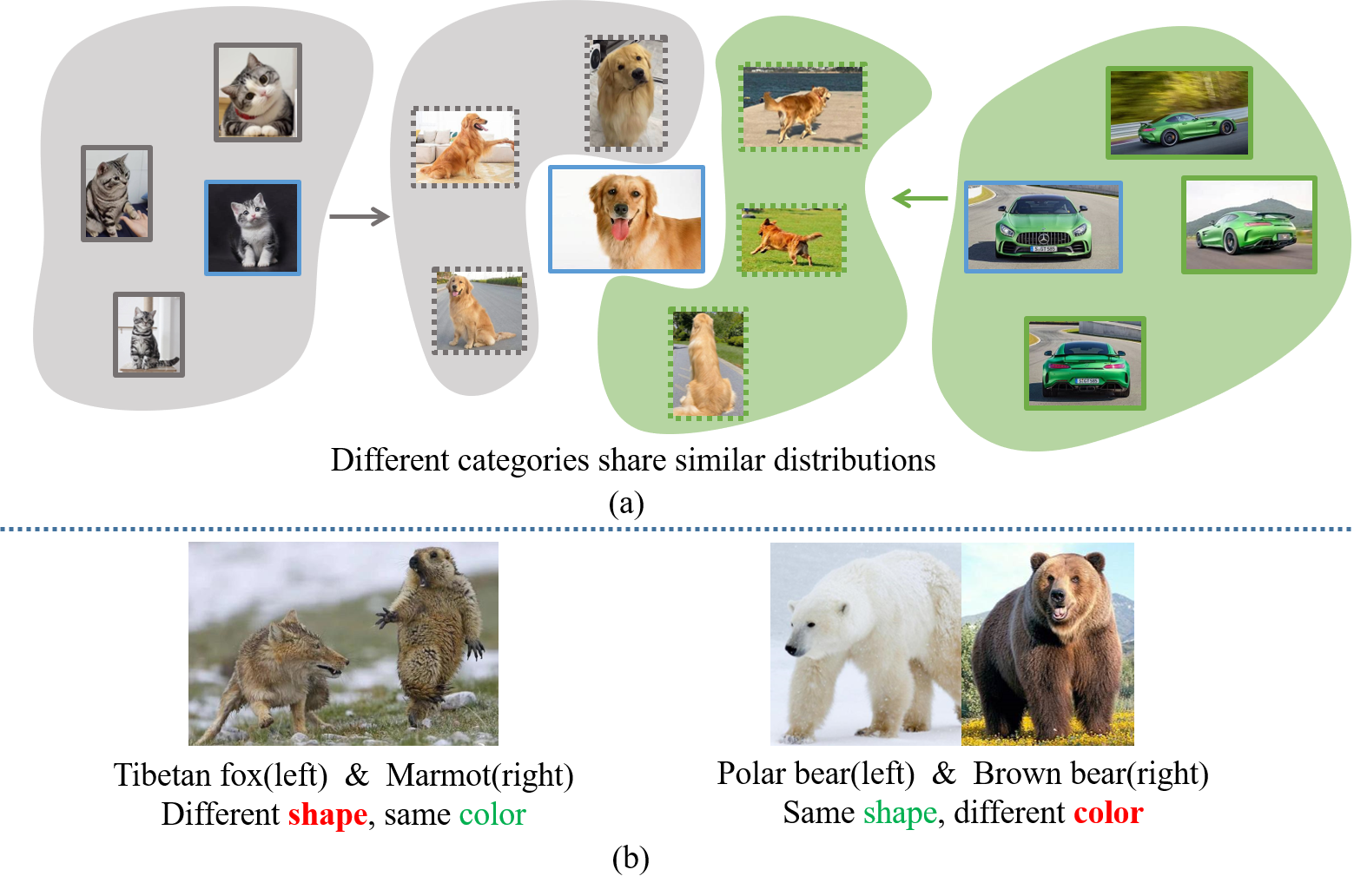}
\end{center}
\caption{The motivation of (a) inter-class analogical augmentation and (b) intra-class significance analysis module.}
\label{fig:motivation}
\end{figure} 

Existing OTFCL methods can be divided into two categories according to whether using a pre-trained model, \emph{i.e.}, train-from-scratch, and -pretrained methods. The train-from-scratch methods~\cite{onlineNoPre1,onlineNoPre2,aljundi2019gradient,onlineNoPre3} follow the representative approaches of continual learning, which focus on preserving old knowledge by devising anti-forgetting loss functions~\cite{onlineNoPre1,onlineNoPre2} or designing memory buffers to store samples ~\cite{aljundi2019gradient,onlineNoPre3}. The train-from-pretrained methods~\cite{ensemble,DSDM} put great emphasis on the structural optimization of the classifier and the establishment of a memory buffer. However, these methods encounter two significant challenges that hinder their performance and applications: 1) they depend on memory buffers storing old samples to mitigate catastrophic forgetting. Nonetheless, this approach is often impractical due to safety and privacy concerns, particularly in personal and institution-related applications. 2) The ability to learn from new samples in a single shot is crucial for OTFCL. Yet, the aforementioned methods only focus on preventing the forgetting of old classes, neglecting the model's capacity to learn from new samples. This limitation diminishes their effectiveness in OTFCL scenarios.

To address the above challenges, we follow train-from-pretrained protocol and propose a novel Non-exemplar (\emph{i.e.}, memory-buffer free) OTFCL method named \textit{Inter-Class Analogical Augmentation and Intra-Class Significance Analysis} (I$^2$CANSAY). I$^2$CANSAY consists of two main modules: 1) an Inter-Class Analogical Augmentation (\textit{ICAN}) module to maintain the knowledge of old classes without requiring memory buffers and 2) an Intra-Class Significance Analysis (\textit{ISAY}) module to efficiently learn the knowledge of new data from one-shot samples. 

More specifically, in \textit{ICAN} module, to get rid of the memory buffers, we define a \emph{prototype} for each class to store the knowledge, where the prototype can be a centroid or weighted mean of the class feature vectors. When learning new classes, we generate \emph{pseudo features} of the old classes using the class prototypes to alleviate catastrophic forgetting. A straightforward pseudo feature generation method is using Gaussian noise~\cite{PASS}. However, the Gaussian assumption does not always hold, resulting in generated pseudo-features lacking diversity. This limitation affects the ability of the generated features to preserve old knowledge effectively. To tackle this problem, we propose generating diverse pseudo features for the old classes according to the feature distributions of different new classes, \emph{i.e.}, inter-class analogical augmentation. The key insights of \textit{ICAN} are two-fold: 1) the feature distribution reveals the diversity of a class and 2) the sample variations of different classes are analogical. For example, in Fig.~\ref{fig:motivation} (a), the old class `dog' shares similar poses such as sitting and running with the new class `cat', and shares similar angle-of-views such as side- and back-view with the new class `car'. In conclusion, the pseudo-features generated by the \textit{ICAN} module encompass diversity across different classes, enabling the retrieval of more valuable information from old classes while learning new ones. This capability better prevents forgetting old classes.

In \textit{ISAY} module, to enhance the ability to learn new data, especially from one-shot samples, we propose a method for analyzing the intra-class attribute significance. This involves measuring the importance of different feature dimensions for classification and correcting classification results by enhancing important dimensions and suppressing irrelevant dimensions. Specifically, when given feature vectors of a class, we first compute the standard deviation (STD) of the features in each dimension. A dimension with a higher STD indicates that the class features fluctuate dramatically on this dimension, revealing that this dimension is less significant for this class, and vice versa. For instance, in Fig.~\ref{fig:motivation} (b), the Tibetan fox and the Marmot are similar in color but different in shape, while the Polar bear and Brown bear are similar in shape but different in color. Therefore, shape features are crucial for recognizing a Marmot, while color features are key for distinguishing a bear. In the implementation, the \textit{ISAY} module generates an importance vector by analyzing the significance of intra-class feature dimensions, serving as a correction bias for the linear classifier. This improves the discriminability of features, effectively enhancing the model's ability to learn knowledge from one-shot samples.

%Therefore,  Strengthening the more significant dimensions and suppressing the irrelevant dimensions, the similarity are more efficient in learning new knowledge from one-shot samples. 
%and calculate the similarity of images to each class as bias to the preliminary classification results accordingly.
%On this basis, we design an intra-class significance analysis method to investigate the feature dimensional significance of different classes.
We conduct comprehensive experiments on four widely used datasets: CIFAR-10, CIFAR-100, CoRe50, and CUB-200, and compare the proposed method with SOTA OTFCL methods and representative continual learning methods. The comparative results indicate that the proposed method outperforms existing approaches, including those utilizing memory buffers, by a significant margin, and achieves SOTA performance. The contributions of our method include:
\begin{itemize}
    \item We have proposed a simple yet effective framework, I$^2$CANSAY, for addressing the OTFCL task. This framework not only effectively prevents catastrophic forgetting without relying on a memory buffer, but also exhibits strong learning capabilities when dealing with online data streams.
    \item We introduce the \textit{ICAN} module for pseudo-sample generation, ensuring authenticity while enhancing the diversity of pseudo-samples. This effectively alleviates forgetting and replaces the need for a memory buffer.
    \item We propose the \textit{ISAY} module to improve the model's learning from one-shot samples. It analyzes feature dimensions' importance for classification, correcting results by enhancing important dimensions and suppressing irrelevant ones.
    \item Extensive experiments on CIFAR-10, CIFAR-100, CoRe50 and CUB-200 demonstrate that our approach achieves state-of-the-art performance in three protocols. 
\end{itemize}
\label{sec:intro}
\vspace{-0.2cm}
\section{Related Work}
\subsection{Offline Continual Learning}
\noindent\textbf{Rehearsal-based} methods involve the model reviewing previous samples during incremental training. For example, iCaRL~\cite{ICARL} and its variants~\cite{ashok2022class,EEIL,lucir,kang2022class} store samples by herding algorithm and design different distillation loss. GAN~\cite{gnn-cl1,gnn-cl2} and PASS~\cite{PASS} generate pseudo samples via prototypes and theoretical distributions. \textbf{Architectural-based} methods focus on modifying the architecture as a key point. PackNet ~\cite{PackNet} and its variants ~\cite{HAT} introduce several strategies to effectively segregate the parameters associated with previous tasks from the current ones. PNN ~\cite{PNN} replicates a new organization for each task, as well as ~\cite{dytox,foster}. \textbf{Regularization-based} methods introduce a regularization term in the loss function. EWC~\cite{EWC} and its variants~\cite{REWC,SI} work on improving the calculation of the parameters' importance. Moreover, ~\cite{LWF,lwm} employs knowledge distillation to align the output of two models on new data. \textbf{Pretrained-based} methods draw inspiration from fine-tuned tasks. FearNet~\cite{fearNet} uses a dual-memory system to consolidate old knowledge, and ILUGAN~\cite{ILUGAN} generates pseudo samples by imitating the feature distribution of stored samples. Current advanced methods, such as L2P \cite{l2p} and its variants \cite{dualprompt,coda}, achieve continual learning by acquiring a small set of insertable prompts for pre-trained ViT models.
\begin{figure*}[htb!]
\setlength{\abovecaptionskip}{-0.05cm} 
\setlength{\belowcaptionskip}{-0.2cm} 
\begin{center}
    \includegraphics[width=0.93\textwidth]{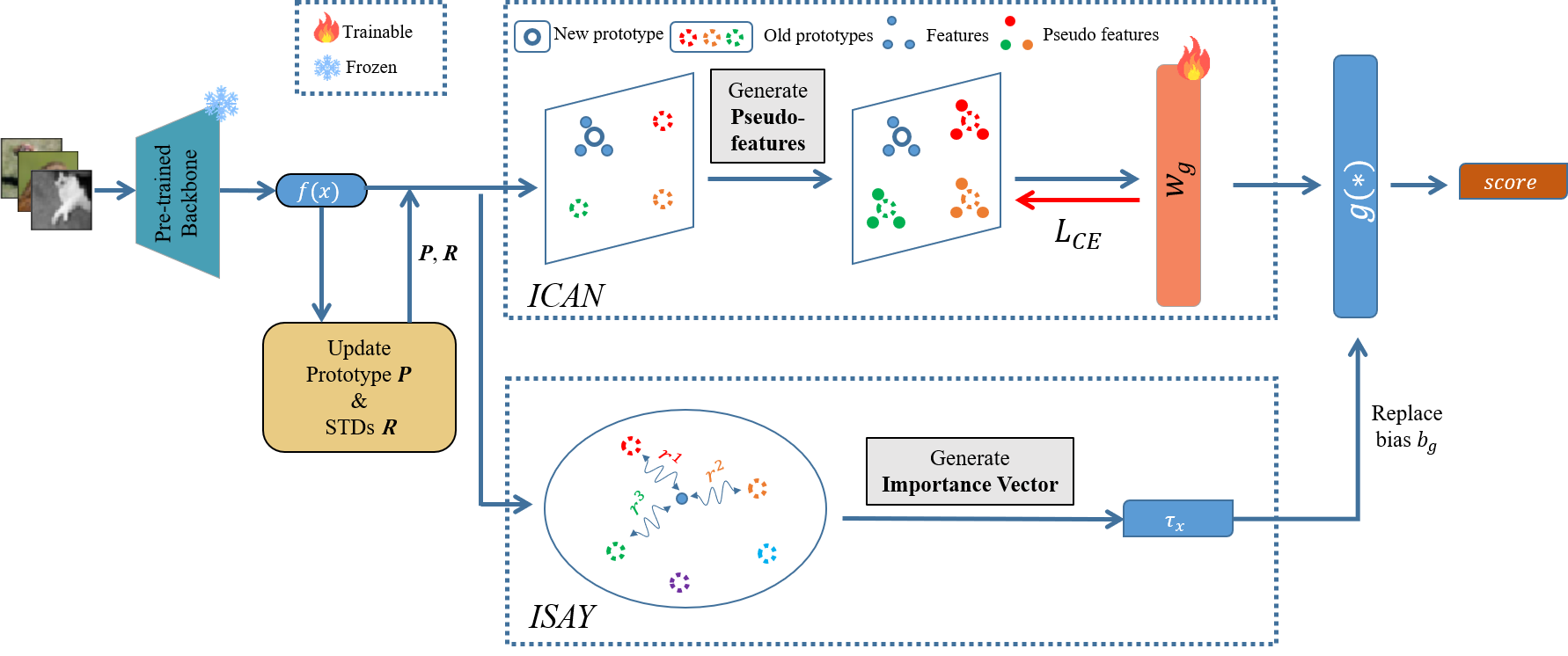}
\end{center}
\caption{The framework of our model. Our model is composed of two modules. The \textit{ICAN} module generates diverse pseudo-features via prototype $P$ and STDs $R$ and trains the classifier's weights $w_g$ by incorporating them along with real features. The \textit{ISAY} module analyzes the intra-class significance attributes for each class via STDs $R$ and generates the importance vector $\tau_x$ as the correction bias for the classifier.}
\label{fig:main}
\end{figure*}
\vspace{-0.25cm}
\subsection{Online Continual Learning}
\noindent\textbf{Online continual learning} involves training the model in an online mode through a continuous stream of input data. There are already many methods~\cite{A-GEM,MIR,voting,GDumb,aljundi2019gradient,ASER} for this task. For instance, ~\cite{aljundi2019gradient} proposes to select samples by the gradient. GDumb~\cite{GDumb} proposes to train the model from scratch using samples only in the memory. MIR~\cite{MIR} trains the model with the maximal interfered retrieval.\\
\textbf{Online task-free continual learning.} Compared to online continual learning, OTFCL abandons the task boundary. Representative OTFCL approaches methods~\cite{onlineNoPre1,onlineNoPre2,onlineNoPre3,DSDM,ensemble,CoPe,CNDPM} can be divided into two categories: 1) Train-from scratch, CoPe ~\cite{CoPe} proposes a framework to determine the stability-plasticity trade-off. GMED~\cite{onlineNoPre3} puts forward to select the representative samples that have a high affinity to past tasks via a gradient. ~\cite{CNDPM} proposes an expansion-based framework to fit the OTFCL protocol. 2) Train-from-pretrained, DSDM~\cite{DSDM} uses the pre-trained backbone and generates a unit pool where each unit consists of a location vector and a label vector. Ensemble~\cite{ensemble} also uses a pre-trained model and trains a single-layer linear corresponding to class labels. Currently, pre-trained models have been widely applied in the field of continual learning. Their strong generalization capabilities make incremental learning algorithms more practically valuable. Therefore, we follow the "train-from-pretrained" protocol and conduct thorough comparisons with relevant SOTA methods.

\vspace{-0.25cm}
\section{Method}
\vspace{-0.1cm}
\subsection{Problem Formulation}

Given an input data stream composed of sequential batches $\mathcal{S}=\{\mathcal{S}^1,...,\mathcal{S}^T\}$, where $T$ represents the total number of batches, $\mathcal{S}^t$ denotes the $t$-th batch. Each batch $\mathcal{S}^t=\{(x_i^t,y_i^t)\}_{i=1}^N$ is composed of $N$ training samples, where $x_i^t\in \mathbb{R}^{H\times W\times C}$ denotes the $i$-th image and $y_i^t\in \mathcal{Y}^t$ denotes its corresponding label. The $\mathcal{Y}^t$ is the label set of $\mathcal{X}^t$. At each incremental learning step $t$, only the current training batch $\mathcal{S}^t$ is available, and the model is expected to recognize all the emerged classes, i.e., $\mathcal{Y}^{1:t}=\mathcal{Y}^1\bigcup\mathcal{Y}^2\bigcup\cdots\bigcup\mathcal{Y}^t$. It is worth mentioning that, in OTFCL, the training samples are one-shot provided, \emph{i.e.}, $\forall m\neq n, \mathcal{S}^m\cap\mathcal{S}^n=\emptyset$, and online learned, \emph{i.e.}, training epoch equals 1.

\vspace{-0.2cm}
\subsection{Framework Overview}
%Fig.\ref{fig:main} shows the diagram of training procedure and testing procedure. Given a 2D image $x\in \mathbb{R}^{H\times W\times C}$, and a neural network $F=f(\ast)\circ g(\ast)$, $f(\ast)$ is a pretrained backbone (without classification head) marked as an encoder, and $g(\ast)$ is the classifier marked as a decoder. During the training procedure, the prototype of each class is stored by the model $P=\left \{ p^0, p^1,\ldots, p^C \right \} $ as well as the variance $R=\left \{ r^0, r^1,\ldots, r^C \right \}$, where $C$ represents the total number of classes the model have seen, $r, p\in \mathbb{R}^D$, and $D$ is the length of feature dimension. The feature of images is extracted by the pretrained encoder $\zeta^x=f(x)$. The decoder is composed of two classifier $g(\ast)=g_{NCM}(\ast) \circ g_{linear}(\ast)$, where $g_{NCM}(\ast)$ decodes the distance between $f^x$ and prototypes $P$ into logical output $distance(f^x, P)\to logits^x_{NCM}$ and it will not be updated during the training sessions because of the fixing of encoder, while $g_{linear}(\ast)$ decodes $\zeta^x$ into logical output directly $\zeta^x\to logits^x_{linear}$ and it is updated by the gradient of classification loss. Given a test image $x^c_t$ at time point $t$, $c$ represents the class this image belongs to.

%, a 2D image $x\in \mathbb{R}^{H\times W\times C}$ is inputted,

Our framework $F$ is composed of an encoder $f(\ast)$ and a decoder $g(\ast)$, $F=g(\ast)\circ f(\ast)$, where $f(\ast)$ is a pre-trained feature extractor (not including the classification head and remaining frozen consistently), and $g(\ast)$ is a linear classifier with parameters comprising weights $w_{g}$ and biases $b_{g}$.

Given an image $x\in \mathbb{R}^{H\times W\times C}$ and its label $y$, the pre-trained encoder $f(\ast)$ first extracts the feature from the inputted image as $f_x \in \mathbb{R}^{D}$, where $D$ represents the length of the feature dimension. Subsequently, the feature $f_x$ updates the prototype $p^{y}$ and dimensional standard deviation (STD) $r^{y}$ for class $y$ through the simple moving average~(SMA) method. The update formula for prototype $p^{y}$ is:
\begin{equation}
\label{sma}
    p^y\leftarrow {\frac{(p^y*n^{y}+f_x)}{n^{y}+1}}, \,\, p^y\in \mathbb{R}^{D} 
\end{equation}
where $*$ is the scalar product and $n^y$ represents the number of images the model has encountered with the label $y$. If the image $x$ is the first instance the model encounters with the label $y$, then the prototype $p^y$ for class $y$ will be directly initialized as $f_x$.

For the update of the standard deviation $r^y$ for class $y$, we first calculate the squared expectation $E(f_x^2)$ for class $y$ and define it as $\hat{p}^y$, where the initialization and update of $\hat{p}^y$ is the same as $p^y$ in Eq.~\eqref{sma}. According to statistical formulas, the STD $r^y$ for class $y$ can be computed using the class mean~(prototype) $p^y$ and its squared expectation $\hat{p}^y$, with the calculation formula as follows:
\begin{equation}
\label{std}
    r^y=\sqrt{D(f_x)}=\sqrt{\hat{p}^y-(p^y)^2}, \,\, r^y\in \mathbb{R}^{D} 
\end{equation}

Therefore, our approach effectively maintains two sets of parameters during incremental learning: the class prototype $P=\left \{ p^1, \ldots, p^N \right \}$ and the squared expectation $\hat{P}=\left \{ \hat{p}^1, \ldots, \hat{p}^N \right \}$ for each class, where N represents all the classes the model has encountered. These parameters are continuously updated through the SMA method as training data flows in. In addition, the standard deviations (STDs) $R=\left \{ r^1, \ldots, r^N \right \}, R\in \mathbb{R}^{N\times D}$ for all classes are calculated in real-time according to Eq.~\eqref{std}.

Subsequently, we input $f_x$ and the updated $P$, $R$ into the \textit{ICAN} module and \textit{ISAY} module, respectively. In particular, the \textit{ICAN} module generates pseudo feature $\zeta_x$ from the aforementioned inputs and trains the weights $w_g$ of the linear classifier $g(\cdot)$ along with the feature $f_x$. The \textit{ICAN} module first computes the distance matrix $dist(f_x, P)$ between the feature $f_x$ and all class prototypes $P$ and calculates the attribute significance matrix $U$ through the standard deviation $R$. Subsequently, generate the importance vector $\tau_{x}$ using $U$ and $dist(f_x, P)$, which serves as the biases $b_{g}$ for the linear classifier $g(\ast)$. The detailed diagram is illustrated in Fig.~\ref{fig:main}.

\begin{figure}[htb]
\setlength{\abovecaptionskip}{-0.1cm} 
\setlength{\belowcaptionskip}{-0.3cm} 
\begin{center}
    \includegraphics[width=0.45\textwidth]{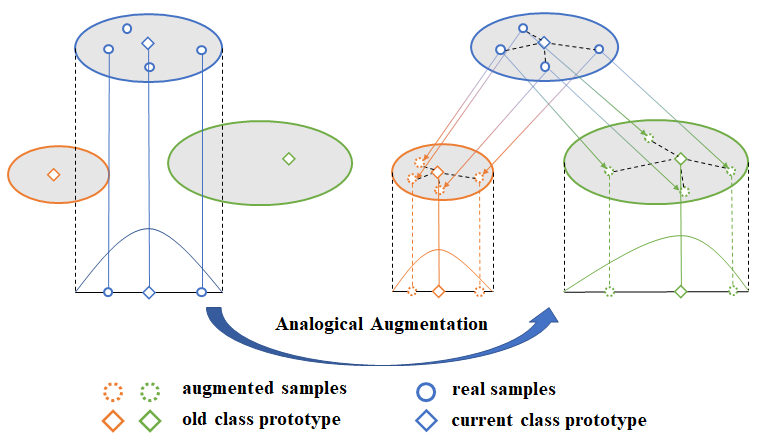}
\end{center}
\caption{The illustration of analogical pseudo features generation. \textit{ICAN} module transfers the feature distribution of new data proportionally to the old classes.}
\label{fig:generation}
\end{figure}
\vspace{-0.1cm}
\subsection{Inter-class Analogical Augmentation}
\label{icaa}
The \textit{ICAN} module is designed to generate pseudo features for old classes based on the input feature. It ensures the authenticity of pseudo-samples through the prototypes of old classes while enhancing their diversity via relative feature distributions. This module effectively alleviates catastrophic forgetting and eliminates the need for memory buffers.

To strike a balance between preventing forgetting and computational burden, we generate only one pseudo-feature corresponding to an old class for each real feature. Specifically, given an image feature $f_x$ and its label $y$, we randomly choose an old class $\bar{y}$ (where $\bar{y} \neq y$) as the label for generating the pseudo-feature. Subsequently, the specific steps for generating the pseudo-feature $\zeta_x$ of the class $\bar{y}$ are as follows. 

First, we calculate the relative distribution vector of real feature $f_x$ through its class prototype $p^{y}$ as $q_x$:  
\begin{equation}
    q_x=f_x-p^{y} , \,\, q_x\in \mathbb{R}^{D} 
\end{equation}

Second, we transfer the relative distribution vector $q_x$ to the old class $\bar{y}$ using the pre-updated $P$ and $R$. Specifically, we normalize $q_x$ by dividing it by its standard deviation $r^y$ to obtain a normalized relative distribution, and then multiply it by $r^{\bar{y}}$ to generate the pseudo distribution.
Finally, the pseudo distribution is added to the class prototype $p^{\bar{y}}$ to obtain the pseudo-feature $\zeta_x$. The function is given by:
\begin{equation}
    \zeta_x=q_x*\frac{r^{\bar{y}}}{r^{y}+\alpha }+p^{\bar{y}}, \,\, \zeta_x\in \mathbb{R}^{D} 
\end{equation}
where $\alpha$ is a small constant to avoid zero division and $*$ is scalar product. The illustration of \textit{ICAN} is shown in Fig.~\ref{fig:generation}.

It's worth noting that our method does not require the use of any distillation loss. This is because the pseudo-features generated by our \textit{ICAN} module can effectively substitute for memory buffers, preventing the linear classifier $g(\ast)$ from forgetting old classes. Moreover, this avoids conflicts between loss functions and the trade-off problem inherent in the distillation loss itself~\cite{topic}. Therefore, we solely use the Cross-Entropy (CE) loss for optimization:
\begin{equation}
   loss=L_{CE}(f_x; w_g)+\beta L_{CE}(\zeta_x;w_g)
\end{equation}
where $\beta$ is the Hyper-parameter. After optimizing the classifier weights $w_g$ to ensure the stability of the old classes, we subsequently adjust the bias $b_g$ to enhance one-shot learning ability.

\begin{figure}[htb]
\setlength{\abovecaptionskip}{-0.1cm} 
\setlength{\belowcaptionskip}{-0.3cm} 
\begin{center}
    \includegraphics[width=0.40\textwidth]{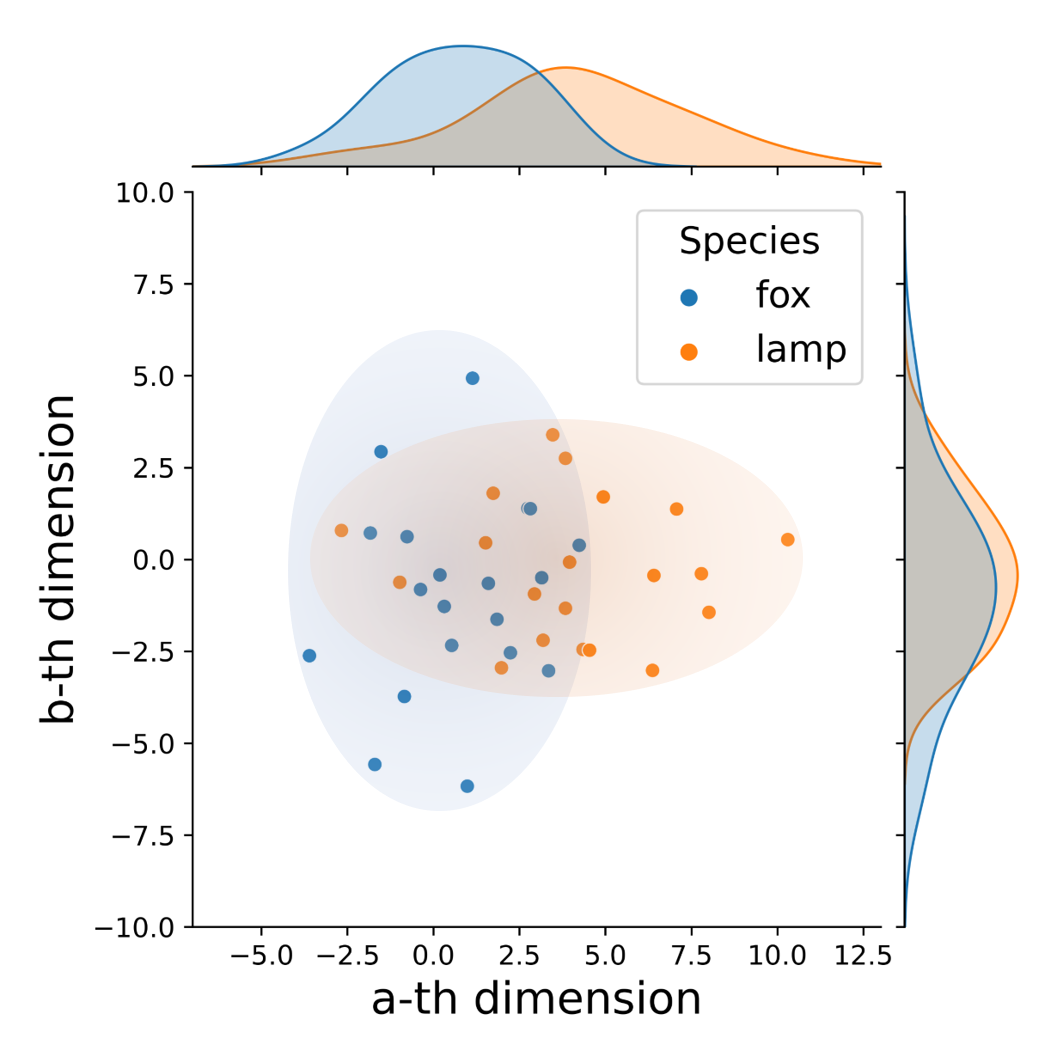}
\end{center}
\caption{The distributions of features in dimensions a and b for the classes `fox' and `lamp', using the dino$\_$ViT8 feature extractor.}
\label{fig:aia}
\end{figure}
\begin{table*}[htb!]
\renewcommand\arraystretch{1.25}
\setlength{\belowcaptionskip}{-0.3cm} 
\scriptsize
\centering
{
\begin{tabular}{l|c|cccc|cccc}
\hline
Dataset &\multirow{2}{*}{Backbone}  &  \multirow{2}{*}{Memory}          & \multicolumn{3}{c|}{CIFAR-10}   &  \multirow{2}{*}{Memory}              & \multicolumn{3}{c}{CIFAR-100}                 \\
Step     &     &     & 2             & 1            & Gaussian   &    & 5            & 1           & Gaussian      \\ \hline
CoPE~\cite{CoPe}    & \multirow{7}{*}{ResNet50}     &   2k   & 48.9          & -             & -  &6k           & -             & 21.6          & -             \\
CN-DPM~\cite{CNDPM}  &       & 2k  & 45.2          & -             & -     &6k        & -             & 20.1          & -             \\
Naive~\cite{ensemble}   &    &    0   & 23.2$\pm$ 4.8   & 10.6$\pm$ 2.2   & 11.4$\pm$ 2.4  & 0 & 3.8$\pm$ 0.5    & 1.0$\pm$0.2    & 6.9$\pm$ 2.9           \\
Ensemble~\cite{ensemble} &      & 2k   & 79.0$\pm$ 0.4   & 78.3$\pm$ 0.4   & 50.1$\pm$ 9.5 & 6k  & 55.3$\pm$ 0.4   & 54.1$\pm$ 0.5   & 39.0$\pm$ 1.4          \\
DSDM, low data~\cite{DSDM} &  & 2k& 79.6$\pm$ 0.5  & 79.4$\pm$ 0.5   & \underline{78.7$\pm$ 1.1}  &6k & 55.3$\pm$ 1.3   & 54.9$\pm$ 1.4   & 55.5$\pm$ 1.2          \\
\textbf{Ours, low data} &  & 0 &\underline{79.7$\pm$ 0.7}         & \underline{79.9$\pm$ 0.4}          & 78.6$\pm$ 0.9       & 0  & \underline{55.7$\pm$0.4}          & \underline{55.5$\pm$0.6}          & \underline{57.1$\pm$0.7}  \\ 
\textbf{Ours} &  & 0 &\textbf{82.9$\pm$ 0.6}         & \textbf{82.5$\pm$ 0.5}          & \textbf{82.8$\pm$ 0.8}        & 0  & \textbf{57.5$\pm$0.5}          & \textbf{57.2$\pm$0.4}          & \textbf{57.3$\pm$0.5}  \\ \hline
L2P~\cite{l2p}   &      \multirow{6}{*}{dino$\_$ViT8}  & 1k& 61.4$\pm$0.7             & 46.8$\pm$0.4             & 57.5$\pm$0.6            & 3k & 27.3$\pm$ 1.2             & 8.4$\pm$0.8             & 47.8$\pm$1.1             \\
DualPrompt~\cite{dualprompt} &  & 1k& 63.7$\pm$0.8             & 49.3$\pm$0.5             & 59.5$\pm$0.8            & 3k & 30.5$\pm$ 1.1          & 10.2$\pm$0.9             & 49.1$\pm$1.1  \\
CODA~\cite{coda} &   &1k & 67.3$\pm$0.7             & 52.7$\pm$0.5             & 65.2$\pm$0.7            & 3k & 33.3$\pm$ 0.9            & 14.3$\pm$0.6             & 56.7$\pm$ 1.0  \\
 DSDM, low data~\cite{DSDM} & &1k & 85.6$\pm$ 0.6          & 85.5$\pm$ 0.7          & 84.9$\pm$ 0.6      &3k   & 60.8$\pm$ 0.9          & 61.1$\pm$ 0.5          & 61.4$\pm$ 1.1          \\
\textbf{Ours, low data} &  & 0& \underline{93.1$\pm$ 0.6} & \underline{92.2$\pm$ 0.5} & \underline{90.9$\pm$ 0.6} & 0 & \underline{71.3$\pm$0.7} & \underline{73.1$\pm$1.1} & \underline{74.0$\pm$1.0} \\
\textbf{Ours}  & & 0& \textbf{93.6$\pm$ 0.8}          & \textbf{93.3$\pm$ 0.6}          & \textbf{91.0$\pm$ 0.8}         & 0 & \textbf{77.2$\pm$0.9}          & \textbf{76.6$\pm$1.4}          &  \textbf{77.3$\pm$1.2}             \\ \hline
\end{tabular}}
\caption{Experiment on OTFCL setting, the mean and stand deviation for 20 runs are as above. Best results are marked in bold; second best results are underlined. The low data means using only $10\%$, $30\%$ data in CIFAR-10 and CIFAR-100 respectively.}
\label{tb:main}
\end{table*}
\vspace{-0.1cm}
\subsection{Intra-class Significance Analysis}
\label{aia}

The feature space distribution is approximated to a normal distribution~\cite{PASS}, but there is a significant difference in the STD across different dimensions. Previous researches suggest that smaller STD in feature dimensions enhances category discriminability, and vice versa. For instance, in Fig.~\ref{fig:aia}, the a-th dimension exhibits a small STD for the fox class, signifying its importance in recognizing 'fox'; similarly, the b-th dimension is more effective in identifying the 'lamp'. Inspired by this, our \textit{ISAY} module adjusts results based on the importance of classification attributes, with detailed calculations as follows.

First, we calculate the attribute significance matrix $U=\left \{ u^1,\ldots,u^N \right \}$ using the standard deviation $R$. As a crucial parameter for \textit{ISAY}, this matrix measures the importance of each dimension across different categories:
\begin{equation}
    u^y=softmax(\max_{i\in [1,D]} (r^y_i)-r^y)
\end{equation}
Where $u^y \in \mathbb{R}^{D}$ represents the attribute importance vector for class $y (y\in[1,N])$ and $r^y_i$ is the STD for $i_{th}$ $ (i\in[1, D])$ dimension.

Then, for input feature $f_x$ with label $y$, we calculate the distance to the prototype $P$, donated as $dist(f_x,P)$. Subsequently, we employ the attribute importance matrix $U$ as a weight to adjust the distance $dist(f_x,P)$, thereby enhancing important dimensions and suppressing less important ones. This process can be formalized as: 
\begin{equation}
    \gamma(f_x, P)=dist(f_x, P)\times U^{\top}, \,\, \gamma(f_x, P)\in \mathbb{R}^{N} 
\end{equation} 
where $dist(\ast)$ represents the euclidean distance and  $\times$ is matrix multiplication. Finally, normalize to generate the importance vector $\tau_x$ based on the adjusted distance:
\begin{equation}
    \tau_x=\left \{ \frac{\left |  \right |   \gamma (f_, P)\left |  \right | _{1} }{\gamma (f_x, P)} \right \}, \,\ \tau_x\in \mathbb{R}^{N}
\end{equation}
where $\left | \left | \cdot  \right |  \right | _{1} $ represents $L1$ norm. 

To convey the importance analysis information to the final result, we employ the importance vector $\tau_x$ as a bias correction, replacing the original bias $b_g$ in the linear classifier $g(\ast)$.
Therefore, the final output of our model is:
\begin{equation}
    F(x)=softmax(f_x \times w_g) + \tau_x
\end{equation}
where $\times$ represents matrix multiplication and the prediction of our model is the $argmax$ of this final output $F(x)$. Moreover, the parameter update of the \textit{ISAY} module is not gradient-based, facilitating the model to acquire more knowledge during the limited online forward and backward propagation, leading to a more effective one-shot learning outcome.

% \clearpage
\section{Experiments}
We benchmark our model with representative OTFCL methods on CIFAR-10 and CIFAR-100. Furthermore, we compare our approach with representative online continual learning methods and offline continual learning methods on CIFAR-10, CoRe50, CIFAR-100, and CUB-200.
\vspace{-0.1cm}
\subsection{Datasets and Experimental Details}
\noindent\textbf{Datasets and benchmark.}
We use (1) Split CIFAR-10~\cite{cifar}, which is composed of 10 classes, with each class containing 5,000 training images and 1,000 testing images. (2) Split CIFAR-100~\cite{cifar}, which contains 100 classes of images, and each class has 500 images for training and 100 for evaluation. (3) CoRe50~\cite{core50} is specifically designed for continual learning which contains 50 classes, around 2400 training images, and 900 testing images for each class. (4) CUB-200~\cite{cub200} includes 200 bird species, and each class has around 30 images for training and 29 images for evaluation. 

\noindent The training data can be partitioned in two ways: first, based on step partition~\cite{DSDM}. For example, Split CIFAR100 (step=1) indicates that each session learns one class, with a total of 100 sessions. Second, employing the Gaussian schedule~\cite{ensemble}, implies that samples from different classes are distributed according to a Gaussian distribution. See the \textbf{Appendix} for more details.\\
\textbf{Evaluation metrics.}
Following the previous methods, we use \textit{Last accuracy ($\%$)} to evaluate models in the OTFCL setting, and \textit{Average accuracy ($\%$)} for offline task-free setting. For the online continual learning setting, we use both metrics.
\begin{table*}[htb!]
\renewcommand\arraystretch{1.25}
\setlength{\belowcaptionskip}{-0.35cm} 
\scriptsize
\centering
{
\begin{tabular}{lcccccccccccc}
\hline
Datasets                   & \multicolumn{6}{c}{CIFAR-10, step=2}                                                 & \multicolumn{6}{c}{CoRe50}                                               \\ \hline
Memory size                & \multicolumn{2}{c}{1k} & \multicolumn{2}{c}{2k} & \multicolumn{2}{c}{5k}           & \multicolumn{2}{c}{1k} & \multicolumn{2}{c}{2k} & \multicolumn{2}{c}{5k} \\ \hline
Accuracy                   & Avg        & Last      & Avg        & Last      & Avg  & \multicolumn{1}{c|}{Last} & Avg        & Last      & Avg        & Last      & Avg        & Last      \\ \hline
\multicolumn{1}{l|}{A-GEM~\cite{A-GEM}} & 43.0       & 17.5      & 59.1       & 38.3      & 74.0 & \multicolumn{1}{c|}{59.0} & 20.7       & 8.4       & 21.9       & 10.3      & 22.9       & 11.5      \\
\multicolumn{1}{l|}{MIR~\cite{MIR}}   & 67.3       & 52.2      & 80.2       & 66.2      & 83.4 & \multicolumn{1}{c|}{74.8} & 33.9       & 21.1      & 37.1       & 24.5      & 38.1       & 27.7      \\
\multicolumn{1}{l|}{GSS~\cite{aljundi2019gradient}}   & 70.3       & 56.7      & 73.6       & 56.3      & 79.3 & \multicolumn{1}{c|}{64.4} & 27.8       & 17.8      & 31.0       & 18.9      & 31.8       & 21.1      \\
\multicolumn{1}{l|}{ASER~\cite{ASER}}  & 63.4       & 46.4      & 78.2       & 59.3      & 83.3 & \multicolumn{1}{c|}{73.1} & 24.3       & 12.2      & 30.8       & 17.4      & 32.5       & 18.5      \\
\multicolumn{1}{l|}{GDUMB~\cite{GDumb}} & 73.8       & 57.7      & 83.8       & 72.4      & 85.3 & \multicolumn{1}{c|}{75.9} & 41.2       & 23.6      & 48.4       & 32.7      & 54.3       & 41.6      \\
\multicolumn{1}{l|}{CV~\cite{voting}}    & 76.0       & 62.9      & 84.9       & 74.1      & 86.1 & \multicolumn{1}{c|}{77.0} & 45.1       & 26.5      & 50.7       & 34.5      & 56.3       & 43.1      \\
\multicolumn{1}{l|}{DSDM~\cite{DSDM}}  & 80.2       & 67.0      & 83.8       & 72.5      & 85.6 & \multicolumn{1}{c|}{76.0} & 43.9       & 43.3      & 53.2       & 50.8      & 66.3       & 57.1      \\ 
\hline
\hline
Memory size                & \multicolumn{12}{c}{\textbf{0}}                                                                                                                                        \\ \hline
Accuracy                   & \multicolumn{3}{c}{Avg}             & \multicolumn{3}{c|}{Last}                    & \multicolumn{3}{c}{Avg}             & \multicolumn{3}{c}{Last}           \\ \hline
\multicolumn{1}{l|}{\textbf{Ours}}  & \multicolumn{3}{c}{\textbf{86.1}}               & \multicolumn{3}{c|}{\textbf{76.6}}                       & \multicolumn{3}{c}{\textbf{71.0}}               & \multicolumn{3}{c}{\textbf{63.2}}              \\ 
\hline
\end{tabular}}
\caption{Comparison results on Split CIFAR-10 and CoRe50 in online continual learning setting. The backbone is ResNet18.}
\label{tb:online}
\end{table*}

\noindent\textbf{Implementation details.}
We train all online models using SGD with $lr=0.02$ and $weight\_ decay=5e-5$. And the pseudo loss weight is set to $\beta =2$, batch size$=50$. We train our offline model with $lr=1e-3$ and $epoch=40$. Our model does \textbf{not} employ memory buffers in any of the comparative experiments, even though other methods use them.

\noindent\textbf{Backbones.} Due to differences in backbone networks across different protocols, we follow previous works by evaluating our model on different pre-trained backbone networks to compare various methods and ensure fairness. Specifically, for OTFCL and offline continual learning settings, we employed a ResNet50 pre-trained on ImageNet as the backbone. To facilitate a broader comparison, the OTFCL setting additionally utilized a dino$\_$ViT8 pre-trained on ImageNet as the backbone. In the case of online continual learning settings, we used a ResNet18 pre-trained on ImageNet as the backbone. All experimental comparison methods use the same pre-trained backbone as our approach for a fair comparison.
%so as CV ~\cite{voting} and DSDM ~\cite{DSDM}, while the other methods do not.
\begin{figure}[htb]
\setlength{\abovecaptionskip}{-0.2cm} 
\setlength{\belowcaptionskip}{-0.4cm} 
\begin{center}
    \includegraphics[width=0.39\textwidth]{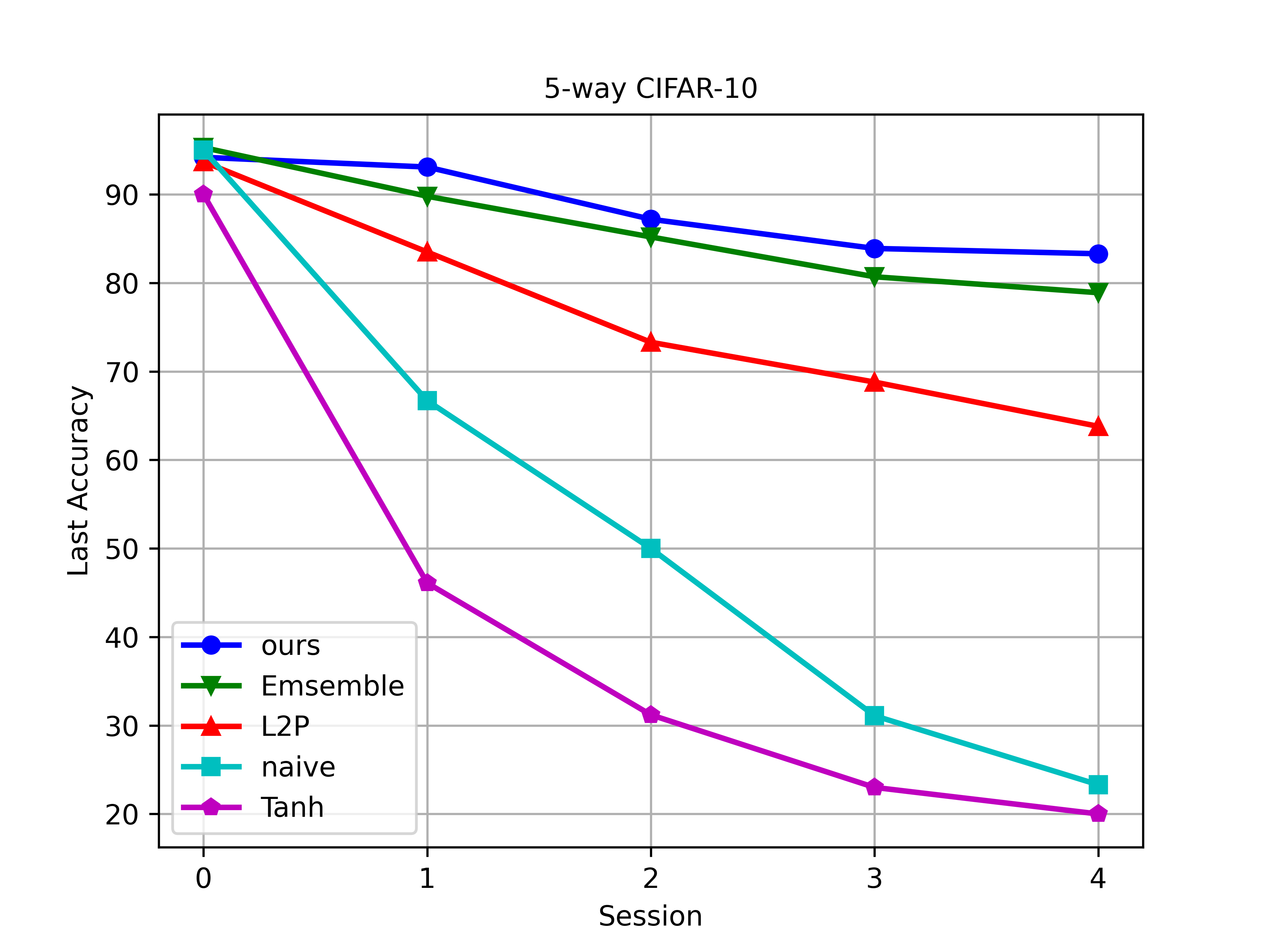}
\end{center}
\caption{The OTFCL session accuracy in Split CIFAR-10 (step=2).}
\label{fig:comparison}
\end{figure}
\vspace{-0.2cm}
\subsection{Online Task-Free Experiments}
\label{main}
In this section, we compare our method on the CIFAR-10 and CIFAR-100 datasets with OTFCL~\cite{CoPe,CNDPM,ensemble,DSDM} methods and SOTA continual learning methods~\cite{l2p,dualprompt,coda}. We present the results of our model on two different backbones: (1) ResNet50, (2) dino$\_$ViT8 \cite{DSDM}. Additionally, we also showcase comparison results under the constraint of a limited-data regime to highlight the online learning capability of our method, where the training data for CIFAR-10 and CIFAR-100 is reduced to $10\%$ and $30\%$, respectively. Note that our method does not use any memory buffer, while other approaches employ a memory buffer with $1$k-$6$k samples. 

\noindent\textbf{Compare Result.}  
As shown in Table~\ref{tb:main}, our method outperforms all the compared methods, reaching SOTA performance. Specifically, on the ResNet50 backbone, we achieve at least a \textbf{3.1\%} and \textbf{2.1\%} improvement over Ensemble on the two datasets, respectively. In the limited-data experiments, our results outperform DSDM by up to \textbf{2.4\%}. Additionally, in the Split CIFAR-10 (step=2), we present the accuracy for each session, and from Fig.~\ref{fig:comparison}, it can be observed that our model consistently outperforms all other methods in each session. On the ViT backbone, our performance advantage is even more pronounced, with an average improvement of \textbf{6.7\%} and \textbf{11.7\%} compared to SOTA methods. 

It's worth noting that (1) even when using a pre-trained model, the Naive method exhibits low accuracy without any forgetting prevention strategy. (2) SOTA continual learning methods L2P, DualPrompt, and CODA perform poorly on the OTFCL protocol, showing a significant gap compared to our method, highlighting the difficulty of the OTFCL task and the effectiveness of our approach.

\label{sec:related}
\begin{table}[htb]
\setlength\tabcolsep{2.8pt}
\setlength{\belowcaptionskip}{-0.3cm} 
\renewcommand\arraystretch{1.25}
\scriptsize
\centering
{
\scriptsize
\begin{tabular}{lll|ll|l}
\hline
Datasets   & \multicolumn{2}{l|}{CIFAR-100} & \multicolumn{2}{l|}{CUB-200} & \multirow{2}{*}{Memory} \\
step        & 5             & 2            & 5           & 2  &       \\ \hline
EWC~\cite{EWC}        & 17.9           & 15.4          & 18.2         & 15.9   & \multirow{7}{*}{1k}  \\
LwF~\cite{LWF}        & 37.9           & 32.8          & 36.1         & 29.8   &      \\
iCaRL~\cite{ICARL}      & 57.2           & 52.8          & 49.6         & 44.3   &      \\
EEIL~\cite{EEIL} & 49.7           & 50.3          & 49.9         & 44.5     &    \\
FearNet~\cite{fearNet}    & 62.5           & 56.9          & 52.7         & 47.8   &      \\
ILUGAN~\cite{ILUGAN}     & 63.1           & 58.0          & 54.9         & 49.7   &      \\
DSDM~\cite{DSDM}       & 63.2           & 63.3          & 55.2         & 55.5   &      \\ \hline
\textbf{Ours}       & \textbf{69.4}               &\textbf{68.2}               &\textbf{64.0}              &\textbf{63.5}    &\textbf{0}          \\ 
\hline
\end{tabular}}
\caption{The comparative results of offline task-free continual learning on the CIFAR-100 and CUB-200 datasets.}
\label{tb:offline}
\end{table}
\vspace{-0.15cm}
\subsection{Other Setting Experiments}
\label{additional}
To verify the generality of our method, we conducted comprehensive experiments under the settings of online continual learning and offline task-free continual learning.\\
\textbf{Online continual learning.}
We compared our method with existing online approaches, including A-GEM~\cite{AGEM}, MIR~\cite{aljundi2019gradient}, GDUMB~\cite{GDumb}, CV~\cite{voting}, GSS~\cite{der+}, and ASER~\cite{ASER}, on the CIFAR-10 and CoRe50 datasets. For CoRe50, following~\cite{DSDM}, we divided it into 9 sessions, with the first session consisting of 10 classes, and the subsequent sessions 5 classes. It is noteworthy that our method operates without any memory buffer, whereas other approaches utilize at least $1$k samples.

In Table~\ref{tb:online}, we observe that existing methods are heavily influenced by the size of the buffer. And our approach still outperforms other methods, even when they use \textbf{5k} buffer samples. Specifically, on the CoRe50 dataset, our method outperforms the second-best method (\textbf{5k} buffer size) by \textbf{4.7\%} and \textbf{6.1\%} in average and Last accuracy. When the buffer size is reduced to \textbf{1k}, our method excels even further, with improvements of \textbf{27.3\%} and \textbf{27.7\%} respectively, demonstrating the robust performance of our approach.\\
\textbf{Offline task-free continual learning.}
We also conducted experiments on CIFAR-100 and CUB-200, comparing with SOTA offline continual learning (CL) methods, including iCaRL~\cite{ICARL}, EEIL~\cite{EEIL}, EWC~\cite{EWC}, FearNet~\cite{fearNet}, LWF~\cite{LWF}, and ILUGAN~\cite{ILUGAN}. In comparison to offline CL, the step partition in the setting of offline task-free CL is shorter (step=$2, 5$), meaning there are more incremental sessions to align with the task-agnostic assumption.

The comparative results are shown in Table~\ref{tb:offline}. Specifically, our model achieves at least \textbf{4.9\%} and \textbf{8.0\%} higher average accuracy on the CIFAR-100 and CUB-200 datasets, respectively, compared to the second-best method.

The results in Table~\ref{tb:online} and~\ref{tb:offline} validate the robust generality of our approach. Our method not only achieves SOTA results in OTCIL tasks but also performs exceptionally well in online CL and offline task-free CL, demonstrating the broad applicability and value of our approach.

\vspace{-0.3cm}
\subsection{Ablation Study}
\label{ablation}
\noindent\textbf{The Effectiveness of Each Component.} Table~\ref{tb:ablation} shows the results of our ablative experiments on Split CIFAR-100 (step=5) in the setting of OTFCL. We built four models to illustrate the contributions of each component to performance improvement. (1) We freeze the backbone and train the classifier jointly with new data and a memory buffer containing $1k$ samples as the baseline. (2) We add an Inter-class Analogical Augmentation (\textit{ICAN}) module to replace the memory buffer. (3) We add
an Intra-class Significance Analysis (\textit{ISAY}) module to improve the baseline model. (4) We add two components to train the model without any memory buffer.
\begin{table}[htb]
\setlength{\belowcaptionskip}{-0.2cm} 
\renewcommand\arraystretch{1.3}
\scriptsize
\centering
\resizebox{0.80\linewidth}{!}{
\setlength{\tabcolsep}{1.6em}
\begin{tabular}{ccc|c|c}
\hline
\multicolumn{3}{c|}{CIFAR-100 (step=5)  ResNet50} & \multirow{2}{*}{Last accuracy (\%)} & \multirow{2}{*}{Memory} \\ \cline{1-3}
   PASS   &         &                       \\ \hline
   Fetr&            &            & 47.2  &     1k                   \\ \hline
      & $\surd$    &            & 58.3          &    0             \\ \hline
  $\surd$     &            & $\surd$    & 57.2         &     1k             \\ \hline
      & $\surd$    & $\surd$    & \textbf{59.4}         &     0              \\ \hline
\end{tabular}}
\caption{The contribution of each component.}
\label{tb:ablation}
\end{table}
As shown in Table~\ref{tb:ablation}, the baseline model produces the lowest last accuracy of \textbf{47.2\%}. Using \textit{ICAN} module not only effectively substitutes for the memory buffer but also results in a relative improvement of \textbf{11.1\%} compared to the baseline. The \textit{ISAY} module significantly enhances the one-shot learning ability, also achieving a substantial improvement of \textbf{10.0\%}. The final model which consists of two components has achieved the best accuracy of \textbf{59.4\%}.

\begin{table}[htb]
\renewcommand\arraystretch{1.25}
\setlength{\belowcaptionskip}{-0.2cm} 
\footnotesize
\centering
\resizebox{0.60\linewidth}{!}{
\setlength{\tabcolsep}{1.0mm}
\begin{tabular}{lllll}
\hline
\multicolumn{5}{l}{CIFAR-100 (step=5)   ResNet50}                     \\ \hline
\multicolumn{1}{l|}{Memory}        & 0    & 1k & 2k & 5k \\ \hline
\multicolumn{1}{l|}{Last accuracy} & 59.4 & 59.2   &59.5    & 59.7   \\ \hline
\end{tabular}}
\caption{The influence of memory size}
\label{tb:memory}
\end{table}
\begin{figure}[htb]
\setlength{\abovecaptionskip}{-0.1cm} 
\setlength{\belowcaptionskip}{-0.2cm} 
\begin{center}
    \includegraphics[width=0.4\textwidth]{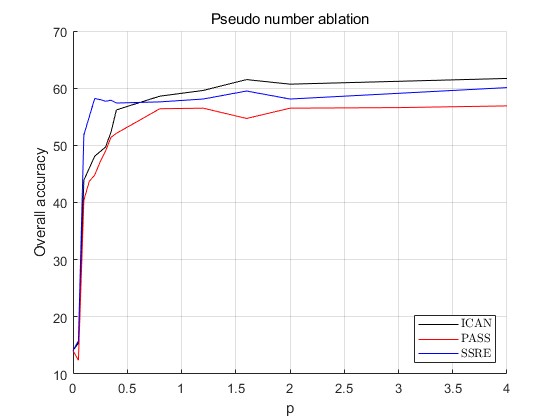}
\end{center}
\caption{The ablation study of the pseudo-feature generation on OTFCL Split CIFAR-100 (step=5) setting. $p$ represents the number of pseudo features divided by the batch size.}
\label{fig:ablation}
\end{figure}
\label{sec:related}
\label{sec:experiment}

\noindent\textbf{The Influence of Buffer Size.} Even though our method does not rely on a memory buffer, for a more comprehensive comparison, we investigated the impact of a memory buffer on our approach on the OTFCL Split CIFAR-100 (step=5) setting. From Table~\ref{tb:memory}, we can observe that as the memory size increases from $0$ to $5$K, the accuracy of our method remains essentially unchanged. This indicates that our approach is not reliant on a memory buffer and further highlights the effectiveness of the proposed pseudo-feature generation module \textit{ICAN}, as it effectively replaces the need for a memory buffer.

\noindent\textbf{Pseudo-feature Generation Analysis.} Fig.~\ref{fig:ablation} shows our ablation study on the generation method and the number of generated pseudo-features. (1) For the generation method, we compare \textit{ICAN} module with Gaussian noise~\cite{PASS} and transferring various distributions~\cite{ssre}. As shown in Fig.~\ref{fig:ablation}, our method achieves better accuracy. (2) As for the generation quantity, we observed that the accuracy rises rapidly with an increase in the number initially, stabilizing when the generation quantity equals the batch size. With further increases in the generation quantity, the accuracy does not show a significant improvement. Considering the balance between training cost and accuracy, we choose the number of generated pseudo-samples to be equal to the batch size, \emph{i.e.}, generating one pseudo-sample for each input feature.

Additional ablation experiments and robustness analysis experiments are provided in \textbf{Appendix}.
\vspace{-0.1cm}
\section{Conclusion}
This paper proposes a novel OTFCL framework, I$^2$CANSAY, which enhances the capability of learning from data streams while eliminating the dependence on memory buffers. The framework comprises two main modules: the \textit{ICAN} module, proven to replace memory buffers by generating highly diverse pseudo-features, achieving superior forgetfulness prevention; the \textit{ISAY} module generates linear correction bias by analyzing intra-class attribute significance, enhancing the ability for online data streams. Comprehensive experiments on three protocols and four datasets demonstrate that our approach achieves SOTA performance without relying on memory buffers, showcasing strong forgetfulness prevention and online learning capabilities.

\label{sec:conclusion}

\appendix
\label{sec:appendix}
\section{Appendix}
\subsection{Implementation Details}
\begin{figure}[h]
\setlength{\belowcaptionskip}{-0.3cm}
\vspace{-0.2cm}
\begin{center}
    \includegraphics[width=0.45\textwidth]{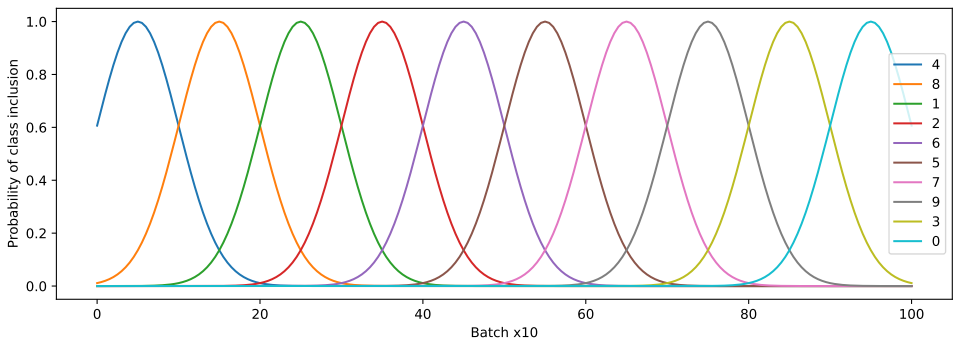}
\end{center}
\vspace{-0.3cm}
\caption{Illustration of data stream when 'step=Gaussian'.} \label{fig:gaussian}
\end{figure}

\noindent\textbf{Date Partitioning.}
 We followed the previous approaches~\cite{DSDM,ensemble} to partition the training data in three protocols, including step partition and Gaussian schedule. Firstly, the term `step' indicates the data division mode. For `step=2', data is categorized, with each of the two categories of samples assigned to a session. Secondly, when step=Gaussian', samples from different classes are distributed following a Gaussian distribution, as illustrated in Figure~\ref{fig:gaussian}.

\noindent\textbf{Setting and Backbone.} According to the experimental settings of the previous method DSDM~\cite{DSDM}, we conducted experiments using three evaluation scenarios, including online task-free continual learning (OTFCL), online continual learning (OCL), and offline task-free continual learning. We followed DSDM~\cite{DSDM}'s fair comparison using different backbones, where OTFCL protocol utilized ResNet50 and dino$\_$ViT8, OCL used ResNet18, and offline task-free continual learning used ResNet50 backbone. We thoroughly compared the respective state-of-the-art methods for each protocol. Additionally, in the OTFCL protocol, we compared with the latest SOTA methods for class-incremental learning, namely L2P~\cite{l2p}, DualPrompt~\cite{dualprompt}, and CODA~\cite{coda}. Since these methods were implemented on the ViT backbone, we re-implemented them on the dino$\_$ViT8 backbone for a fair comparison (only changing the backbone and replicating according to their codebase and hyperparameters).

\subsection{More Ablation Study}
\noindent\textbf{Sensitive Study of Hyper-parameter $\beta$.} We conducted a sensitivity analysis on the hyper-parameter $\beta$ in the OTFCL Split CIFAR-10 (step=1) experiment. Specifically, we tested $\beta$ values of 0.5, 1.0, 1.5, 2.0, 2.5, and 3.0, respectively. From Table~\ref{tb:111}, we observe that with the increase of $\beta$, the accuracy first increases and then decreases. The optimal performance is attained when $\beta=2$, yielding an accuracy of 82.73\%. Moreover, as $\beta$ changes, the fluctuation in accuracy is minimal, demonstrating the robustness of our method.

\begin{table}[htb]
\renewcommand\arraystretch{1.25}
\setlength{\belowcaptionskip}{-0.2cm} 
\footnotesize
\centering
\resizebox{0.8\linewidth}{!}{
\setlength{\tabcolsep}{1.0mm}
\begin{tabular}{lcccccc}
\hline
\multicolumn{7}{c}{CIFAR-10 (step=1) ResNet50}                     \\ \hline
\multicolumn{1}{l|}{$\beta$}        & 0.5   & 1.0 & 1.5 & 2 & 2.5 & 3.0\\ \hline
\multicolumn{1}{l|}{Last accuracy (\%)} & 82.12 & 82.43 & 82.61   &  82.73   & 82.67 & 82.65    \\ \hline
\end{tabular}}
\caption{Sensitive study of hyper-parameter $\beta$.}
\label{tb:111}
\end{table}

{\small
\bibliographystyle{ieee_fullname}
\bibliography{egbib}

\begin{thebibliography}{10}\itemsep=-1pt

\bibitem{MIR}
Rahaf Aljundi, Eugene Belilovsky, Tinne Tuytelaars, Laurent Charlin, Massimo Caccia, Min Lin, and Lucas Page-Caccia.
\newblock Online continual learning with maximal interfered retrieval.
\newblock In H. Wallach, H. Larochelle, A. Beygelzimer, F. d\textquotesingle Alch\'{e}-Buc, E. Fox, and R. Garnett, editors, {\em Advances in Neural Information Processing Systems 32}, pages 11849--11860. Curran Associates, Inc., 2019.

\bibitem{taskfree}
Rahaf Aljundi, Klaas Kelchtermans, and Tinne Tuytelaars.
\newblock Task-free continual learning.
\newblock In {\em IEEE Conference on Computer Vision and Pattern Recognition}, pages 11254--11263, 2019.

\bibitem{aljundi2019gradient}
Rahaf Aljundi, Min Lin, Baptiste Goujaud, and Yoshua Bengio.
\newblock Gradient based sample selection for online continual learning.
\newblock {\em Advances in neural information processing systems}, 32, 2019.

\bibitem{ashok2022class}
Arjun Ashok, KJ Joseph, and Vineeth~N Balasubramanian.
\newblock Class-incremental learning with cross-space clustering and controlled transfer.
\newblock In {\em European Conference on Computer Vision}, pages 105--122. Springer, 2022.

\bibitem{der+}
Pietro Buzzega, Matteo Boschini, Angelo Porrello, Davide Abati, and Simone Calderara.
\newblock Dark experience for general continual learning: a strong, simple baseline.
\newblock {\em Advances in neural information processing systems}, 33:15920--15930, 2020.

\bibitem{EEIL}
Francisco~M Castro, Manuel~J Mar{\'\i}n-Jim{\'e}nez, Nicol{\'a}s Guil, Cordelia Schmid, and Karteek Alahari.
\newblock End-to-end incremental learning.
\newblock In {\em European Conference on Computer Vision}, pages 233--248, 2018.

\bibitem{A-GEM}
Arslan Chaudhry, Marc'Aurelio Ranzato, Marcus Rohrbach, and Mohamed Elhoseiny.
\newblock Efficient lifelong learning with a-gem.
\newblock {\em arXiv preprint arXiv:1812.00420}, 2018.

\bibitem{AGEM}
Arslan Chaudhry, Marc'Aurelio Ranzato, Marcus Rohrbach, and Mohamed Elhoseiny.
\newblock Efficient lifelong learning with a-gem.
\newblock {\em arXiv preprint arXiv:1812.00420}, 2018.

\bibitem{CoPe}
Matthias De~Lange, Rahaf Aljundi, Marc Masana, Sarah Parisot, Xu Jia, Ale{\v{s}} Leonardis, Gregory Slabaugh, and Tinne Tuytelaars.
\newblock A continual learning survey: Defying forgetting in classification tasks.
\newblock {\em IEEE transactions on pattern analysis and machine intelligence}, 44(7):3366--3385, 2021.

\bibitem{lwm}
Prithviraj Dhar, Rajat~Vikram Singh, Kuan-Chuan Peng, Ziyan Wu, and Rama Chellappa.
\newblock Learning without memorizing.
\newblock In {\em IEEE/CVF Conference on Computer Vision and Pattern Recognition}, pages 5138--5146, 2019.

\bibitem{dong2021few}
Songlin Dong, Xiaopeng Hong, Xiaoyu Tao, Xinyuan Chang, Xing Wei, and Yihong Gong.
\newblock Few-shot class-incremental learning via relation knowledge distillation.
\newblock In {\em AAAI Conference on Artificial Intelligence}, volume~35, pages 1255--1263, 2021.

\bibitem{dytox}
Arthur Douillard, Alexandre Ram{\'e}, Guillaume Couairon, and Matthieu Cord.
\newblock Dytox: Transformers for continual learning with dynamic token expansion.
\newblock In {\em IEEE/CVF Conference on Computer Vision and Pattern Recognition}, pages 9285--9295, 2022.

\bibitem{voting}
Jiangpeng He and Fengqing Zhu.
\newblock Online continual learning via candidates voting.
\newblock In {\em Proceedings of the IEEE/CVF winter conference on applications of computer vision}, pages 3154--3163, 2022.

\bibitem{onlineNoPre1}
Yong~Won Hong, Hyeran Byun, and Sungho Park.
\newblock Task-aware network: Mitigation of task-aware and task-free performance gap in online continual learning.
\newblock {\em Available at SSRN 4342057}.

\bibitem{lucir}
Saihui Hou, Xinyu Pan, Chen~Change Loy, Zilei Wang, and Dahua Lin.
\newblock Learning a unified classifier incrementally via rebalancing.
\newblock In {\em IEEE Conference on Computer Vision and Pattern Recognition}, pages 831--839, 2019.

\bibitem{onlineNoPre3}
Xisen Jin, Arka Sadhu, Junyi Du, and Xiang Ren.
\newblock Gradient-based editing of memory examples for online task-free continual learning.
\newblock {\em Advances in Neural Information Processing Systems}, 34:29193--29205, 2021.

\bibitem{kang2022class}
Minsoo Kang, Jaeyoo Park, and Bohyung Han.
\newblock Class-incremental learning by knowledge distillation with adaptive feature consolidation.
\newblock In {\em IEEE/CVF Conference on Computer Vision and Pattern Recognition}, pages 16071--16080, 2022.

\bibitem{fearNet}
Ronald Kemker and Christopher Kanan.
\newblock Fearnet: Brain-inspired model for incremental learning.
\newblock {\em arXiv preprint arXiv:1711.10563}, 2017.

\bibitem{EWC}
James Kirkpatrick, Razvan Pascanu, Neil Rabinowitz, Joel Veness, Guillaume Desjardins, Andrei~A Rusu, Kieran Milan, John Quan, Tiago Ramalho, Agnieszka Grabska-Barwinska, et~al.
\newblock Overcoming catastrophic forgetting in neural networks.
\newblock {\em National Academy of Sciences}, 114(13):3521--3526, 2017.

\bibitem{cifar}
Alex Krizhevsky and Geoffrey Hinton.
\newblock Learning multiple layers of features from tiny images.
\newblock Technical report, Citeseer, 2009.

\bibitem{CNDPM}
Soochan Lee, Junsoo Ha, Dongsu Zhang, and Gunhee Kim.
\newblock A neural dirichlet process mixture model for task-free continual learning.
\newblock {\em arXiv preprint arXiv:2001.00689}, 2020.

\bibitem{LWF}
Zhizhong Li and Derek Hoiem.
\newblock Learning without forgetting.
\newblock {\em IEEE Transactions on Pattern Analysis and Machine Intelligence}, 40(12):2935--2947, 2017.

\bibitem{REWC}
Xialei Liu, Marc Masana, Luis Herranz, Joost Van~de Weijer, Antonio~M Lopez, and Andrew~D Bagdanov.
\newblock Rotate your networks: Better weight consolidation and less catastrophic forgetting.
\newblock In {\em International Conference on Pattern Recognition}, pages 2262--2268. IEEE, 2018.

\bibitem{core50}
Vincenzo Lomonaco and Davide Maltoni.
\newblock Core50: a new dataset and benchmark for continuous object recognition.
\newblock In {\em Conference on Robot Learning}, pages 17--26. PMLR, 2017.

\bibitem{mai2022online}
Zheda Mai, Ruiwen Li, Jihwan Jeong, David Quispe, Hyunwoo Kim, and Scott Sanner.
\newblock Online continual learning in image classification: An empirical survey.
\newblock {\em Neurocomputing}, 469:28--51, 2022.

\bibitem{PackNet}
Arun Mallya and Svetlana Lazebnik.
\newblock Packnet: Adding multiple tasks to a single network by iterative pruning.
\newblock In {\em IEEE Conference on Computer Vision and Pattern Recognition}, pages 7765--7773, 2018.

\bibitem{DSDM}
Julien Pourcel, Ngoc-Son Vu, and Robert~M French.
\newblock Online task-free continual learning with dynamic sparse distributed memory.
\newblock In {\em Computer Vision--ECCV 2022: 17th European Conference, Tel Aviv, Israel, October 23--27, 2022, Proceedings, Part XXV}, pages 739--756. Springer, 2022.

\bibitem{GDumb}
Ameya Prabhu, Philip~HS Torr, and Puneet~K Dokania.
\newblock Gdumb: A simple approach that questions our progress in continual learning.
\newblock In {\em Computer Vision--ECCV 2020: 16th European Conference, Glasgow, UK, August 23--28, 2020, Proceedings, Part II 16}, pages 524--540. Springer, 2020.

\bibitem{gnn-cl2}
Jason Ramapuram, Magda Gregorova, and Alexandros Kalousis.
\newblock Lifelong generative modeling.
\newblock {\em Neurocomputing}, 404:381--400, 2020.

\bibitem{ICARL}
Sylvestre-Alvise Rebuffi, Alexander Kolesnikov, Georg Sperl, and Christoph~H Lampert.
\newblock icarl: Incremental classifier and representation learning.
\newblock In {\em IEEE/CVF Conference on Computer Vision and Pattern Recognition}, pages 2001--2010, 2017.

\bibitem{PNN}
Andrei~A Rusu, Neil~C Rabinowitz, Guillaume Desjardins, Hubert Soyer, James Kirkpatrick, Koray Kavukcuoglu, Razvan Pascanu, and Raia Hadsell.
\newblock Progressive neural networks.
\newblock {\em arXiv preprint arXiv:1606.04671}, 2016.

\bibitem{HAT}
Joan Serra, Didac Suris, Marius Miron, and Alexandros Karatzoglou.
\newblock Overcoming catastrophic forgetting with hard attention to the task.
\newblock In {\em International Conference on Machine Learning}, pages 4548--4557. PMLR, 2018.

\bibitem{ensemble}
Murray Shanahan, Christos Kaplanis, and Jovana Mitrovi{\'c}.
\newblock Encoders and ensembles for task-free continual learning.
\newblock {\em arXiv preprint arXiv:2105.13327}, 2021.

\bibitem{ASER}
Dongsub Shim, Zheda Mai, Jihwan Jeong, Scott Sanner, Hyunwoo Kim, and Jongseong Jang.
\newblock Online class-incremental continual learning with adversarial shapley value.
\newblock In {\em Proceedings of the AAAI Conference on Artificial Intelligence}, volume~35, pages 9630--9638, 2021.

\bibitem{gnn-cl1}
Hanul Shin, Jung~Kwon Lee, Jaehong Kim, and Jiwon Kim.
\newblock Continual learning with deep generative replay.
\newblock {\em Neural Information Processing Systems}, 30, 2017.

\bibitem{coda}
James~Seale Smith, Leonid Karlinsky, Vyshnavi Gutta, Paola Cascante-Bonilla, Donghyun Kim, and et al.
\newblock Coda-prompt: Continual decomposed attention-based prompting for rehearsal-free continual learning.
\newblock In {\em Proceedings of the IEEE/CVF Conference on Computer Vision and Pattern Recognition}, pages 11909--11919, 2023.

\bibitem{topic}
Xiaoyu Tao, Xiaopeng Hong, Xinyuan Chang, Songlin Dong, Xing Wei, and Yihong Gong.
\newblock Few-shot class-incremental learning.
\newblock In {\em IEEE/CVF Conference on Computer Vision and Pattern Recognition}, pages 12183--12192, 2020.

\bibitem{foster}
Fu-Yun Wang, Da-Wei Zhou, Han-Jia Ye, and De-Chuan Zhan.
\newblock Foster: Feature boosting and compression for class-incremental learning.
\newblock {\em arXiv preprint arXiv:2204.04662}, 2022.

\bibitem{dualprompt}
Zifeng Wang, Zizhao Zhang, Sayna Ebrahimi, Ruoxi Sun, Han Zhang, Chen-Yu Lee, Xiaoqi Ren, Guolong Su, Vincent Perot, Jennifer Dy, et~al.
\newblock Dualprompt: Complementary prompting for rehearsal-free continual learning.
\newblock In {\em European Conference on Computer Vision}, pages 631--648. Springer, 2022.

\bibitem{l2p}
Zifeng Wang, Zizhao Zhang, Chen-Yu Lee, Han Zhang, Ruoxi Sun, Xiaoqi Ren, Guolong Su, Vincent Perot, Jennifer Dy, and Tomas Pfister.
\newblock Learning to prompt for continual learning.
\newblock In {\em Proceedings of the IEEE/CVF Conference on Computer Vision and Pattern Recognition}, pages 139--149, 2022.

\bibitem{cub200}
Peter Welinder, Steve Branson, Takeshi Mita, Catherine Wah, Florian Schroff, Serge Belongie, and Pietro Perona.
\newblock Caltech-ucsd birds 200.
\newblock 2010.

\bibitem{ILUGAN}
Ye Xiang, Ying Fu, Pan Ji, and Hua Huang.
\newblock Incremental learning using conditional adversarial networks.
\newblock In {\em Proceedings of the IEEE/CVF International Conference on Computer Vision}, pages 6619--6628, 2019.

\bibitem{onlineNoPre2}
Fei Ye and Adrian~G Bors.
\newblock Learning dynamic latent spaces for lifelong generative modelling.
\newblock 2023.

\bibitem{SI}
Friedemann Zenke, Ben Poole, and Surya Ganguli.
\newblock Continual learning through synaptic intelligence.
\newblock In {\em International Conference on Machine Learning}, pages 3987--3995. JMLR. org, 2017.

\bibitem{PASS}
Fei Zhu, Xu-Yao Zhang, Chuang Wang, Fei Yin, and Cheng-Lin Liu.
\newblock Prototype augmentation and self-supervision for incremental learning.
\newblock In {\em Proceedings of the IEEE/CVF Conference on Computer Vision and Pattern Recognition}, pages 5871--5880, 2021.

\bibitem{ssre}
Kai Zhu, Wei Zhai, Yang Cao, Jiebo Luo, and Zheng-Jun Zha.
\newblock Self-sustaining representation expansion for non-exemplar class-incremental learning.
\newblock In {\em Proceedings of the IEEE/CVF Conference on Computer Vision and Pattern Recognition}, pages 9296--9305, 2022.

\end{thebibliography}
}

\end{document}